%% file: template.tex
\documentclass[a4paper]{article}
\usepackage{INTERSPEECH2021}
\usepackage{multicol}
\usepackage{graphicx}
\usepackage{amsmath}
\usepackage{kotex}
\usepackage{times}
\usepackage{latexsym}
\usepackage{mathtools}
\usepackage{multirow}
\usepackage{adjustbox}
\usepackage{color,soul}
\usepackage{microtype}
\usepackage{adjustbox}
\usepackage{latexsym}
\usepackage{multirow}
\usepackage{diagbox}
\usepackage{amsmath}
\usepackage{pifont}
\usepackage{graphicx}
\usepackage{amssymb}
\usepackage{arydshln}

\usepackage{booktabs}
\usepackage{caption}
\usepackage{subcaption}
\usepackage{color}
\usepackage{makecell}
\usepackage{xcolor}
\usepackage{algorithm}
\usepackage{algpseudocode}
\usepackage{pifont}
\usepackage{amsmath}
\usepackage{xcolor}
\definecolor{cusorange}{RGB}{255, 50, 50}
\definecolor{medblue}{RGB}{28, 98, 215}
\definecolor{darkgreen}{RGB}{0, 100, 0}
\definecolor{darkred}{RGB}{167, 51, 51}

\title{Auxiliary Sequence Labeling Tasks for Disfluency Detection}
\name{
\begin{tabular}{c}
Dongyub Lee\textsuperscript{1}, Byeongil Ko\textsuperscript{2}, Myeong Cheol Shin\textsuperscript{2}, Taesun Whang\textsuperscript{3}, \\ Daniel Lee\textsuperscript{2}, Eunhwa Kim\textsuperscript{2}, Eunggyun Kim\textsuperscript{2}, and Jaechoon Jo\textsuperscript{4}
\end{tabular}
}
\address{\textsuperscript{1}Kakao Corp.\\\textsuperscript{2}Kakao Enterprise\\\textsuperscript{3}Wisenut Inc.\\\textsuperscript{4}Hanshin University}
\email{jude.lee@kakaocorp.com, \{kobi.k,index.sh,daniel.e,silvia.k,jason.ng\}@kakaoenterprise.com, taesunwhang@gmail.com, jaechoon@hs.ac.kr}

\begin{document}

\maketitle

\input{sections/0_abstract}
\input{sections/1_introduction}
\input{sections/2_related_work}

\input{sections/3_model}
\input{sections/4_experiment}

\input{sections/5_conclusion}

\bibliographystyle{IEEEtran}

\bibliography{mybib}


\end{document}

%% file: sections/0_abstract.tex
\begin{abstract}
Detecting disfluencies in spontaneous speech is an important preprocessing step in natural language processing and speech recognition applications. Existing works for disfluency detection have focused on designing a single objective only for disfluency detection, while auxiliary objectives utilizing linguistic information of a word such as named entity or part-of-speech information can be effective. In this paper, we focus on detecting disfluencies on spoken transcripts and propose a method utilizing named entity recognition (NER) and part-of-speech (POS) as auxiliary sequence labeling (SL) tasks for disfluency detection. First, we investigate cases that utilizing linguistic information of a word can prevent mispredicting important words and can be helpful for the correct detection of disfluencies. Second, we show that training a disfluency detection model with auxiliary SL tasks can improve its F-score in disfluency detection. Then, we analyze which auxiliary SL tasks are influential depending on baseline models. Experimental results on the widely used English Switchboard dataset show that our method outperforms the previous state-of-the-art in disfluency detection.
\end{abstract}

%% file: sections/1_introduction.tex
\section{Introduction}
\label{sec:intro}
Detecting disfluencies in spontaneous speech presents a significant challenge for natural language processing (NLP) problems such as parsing and machine translation~\cite{honnibal2014joint,wang2010automatic}. Speech disfluencies refer to any pauses in the normal flow of speech, including corrections, false starts, filled pauses, and repetitions. Figure~\ref{fig:example_1} represents a standard annotation of disfluency structure. Three discrete parts of disfluency annotations: reparandum, interregnum, and repair are defined in~\cite{shriberg1994preliminaries}. The reparandum (RM) indicates words that the speaker intends to discard, interruption point (+) indicates the end of reparandum, interregnum (IM) indicates such as ﬁlled pauses, and discourse cue words and repair (RP) indicates correct words. 

\begin{figure}[h]\centering
\includegraphics[width=0.4\textwidth]{imgs/fig_1.pdf}
\caption{An example of standard disfluency annotation in the English Switchboard dataset.}
\label{fig:example_1}
\end{figure}

In general, interregnums are relatively easy to detect than reparandums because they have fixed phrases (e.g. ``um", ``uh"). As a result, the main challenge is detecting reparandums which having a free form of structure. As with most previous researches, we also focus on detecting reparandums. In disfluency detection, it is crucial not only to increase recall but also to reduce the case of predicting fluent speech is not fluent. 

On the one hand, words representing a disfluency tend to be meaningless. On the other hand, words representing named entities are usually meaningful. For example, they can refer to locations, time, and names of people. Table~\ref{table:intro} shows that utilizing named entities can be highly effective in the disfluency detection task. 
In Table~\ref{table:intro}, the incorrectly predicted words ``my forty" correspond to a named entity (TIME) while the ground-truth disfluencies are not named entities. Therefore, we assume that utilizing named entity recognition (NER) in the disfluency detection task can prevent mispredicting important words classified as named entities as disfluency, and eventually will improve the performance of disfluency detection.
\input{tables/tab0_intro}


\begin{figure}[h]\centering
\includegraphics[width=0.48\textwidth]{imgs/fig_2.pdf}
\caption{Examples of cases where RM and RP have the same POS tag type. The right side marked with / indicates the pos tag type.
}
\label{fig:example_2}
\end{figure}

Figure~\ref{fig:example_2} represents the cases where RM and RP have the same POS tag type. Since the same POS tag sequence is used repeatedly, we hypothesize that if the POS tag is predicted at the same time during the disfluency prediction, it can lead to improved performance of disfluency detection. We also noticed that as a result of examining the pos tag distribution of words constituting the disfluencies, the ratio of `, (comma)' and `prp (personal pronoun)' tags accounts for 55\%. These two observations led us to use the POS prediction task for the disfluency detection task.

In this paper, we propose a method to increase the performance of disfluency detection by leveraging NER and POS tagging tasks, which are representative sequence labeling tasks. Although many approaches have been proposed for disfluency detection, most of the works only leverage disfluency label as target label~\cite{wang2017transition, lou2018disfluency, zayats2019giving, bach2019noisy}. \cite{snover-etal-2004-lexically} pointed out leveraging POS as a feature can be beneficial and utilized it as a rule-based method.  However rule-based approaches have a limitation in that it can not reflect the deep semantic meaning of words in sentences.  \cite{zayats2016disfluency} utilize POS information as a feature-based method. However a feature-based method requires extra time costs in inference time because all features have to be extracted. Since our method only utilizes NER and POS auxiliary tasks only in training time, our work does not require additional computational costs in inference time. Our work is similar to~\cite{wang2019multi} but differs in that they leverage self-supervised learning to solve the bottleneck of training data, and we leverage sequence labeling tasks as auxiliary tasks for disfluency detection.

To the best of our knowledge, our work is the first attempt to utilizes NER and POS as a multi-task learning approach in disfluency detection. Our contributions can be summarized as follows:
\begin{itemize}
\item We propose the joint training method utilizing NER and POS sequence labeling tasks as auxiliary tasks, which allows the shared encoder to learn a deep semantic representation of words, also reflecting NER and POS meaning representation for disfluency detection.
\item Through quantitative and qualitative analysis on the widely used English Switchboard dataset, we demonstrate that the proposed method can lead to better performance of disfluency detection than not utilizing NER and POS auxiliary tasks.
\item The proposed method achieved state-of-the-art result over previous works on the English Switchboard dataset.
\end{itemize}

%% file: tables/tab0_intro.tex

\begin{table}[t] \centering
\renewcommand{\arraystretch}{1.1}
\begin{adjustbox}{width=0.48\textwidth}
\resizebox{1.0\columnwidth}{!}{
\begin{tabular}{lccc}
\toprule

Examples   & Model & Predictions   \\
\midrule
\multirow{3}{*}{\begin{tabular}[c]{@{}l@{}}
And \textcolor{medblue}{i would,} once i start my forty\textcolor{darkgreen}{$($B-TIME$)$},\\ i'd like to do the forty-five\textcolor{darkgreen}{$($B-TIME$)$} minutes\textcolor{darkgreen}{$($I$)$}\\ a\textcolor{darkgreen}{$($B-DATE$)$} day\textcolor{darkgreen}{$($I$)$} on the bike for a week.\end{tabular}}
& \multirow{2}{*}{\begin{tabular}[c]{@{}c@{}} Ours \\ w/o aux\end{tabular}} 
& \multirow{2}{*}{\begin{tabular}[c]{@{}c@{}} my \\forty\end{tabular}}   \\
&    &     &     \\

\cdashline{2-4}
& Ours  & i would,   \\

\bottomrule

\end{tabular}
}
\end{adjustbox}
\caption{Words painted in blue represent ground-truth disfluencies. Named entities following begin-inside-outside (BIO) scheme are painted in green. Ours w/o aux denotes the model trained without auxiliary SL tasks. 
}
\label{table:intro}
\vspace{-0.5cm}

\end{table}

%% file: sections/2_related_work.tex
\section{Related Work}
\label{sec:related_work}
\noindent \textbf{Sequence Labeling Tasks} \indent Representative sequence labeling tasks include NER and POS tagging tasks. NER refers to the task of predicting entities and POS refers to predicting corresponding part of a speech tag, based on its context. Recent works for both tasks utilize convolutional or recurrent neural networks, transformer, and BERT with CRF layer~\cite{devlin2018bert, lee2018character}.

\noindent \textbf{Disfluency Detection} \indent Disfluency detection is commonly classified into parsing-based~\cite{zwarts2011impact}, translation-based~\cite{wang2017transition, wang2018semi}, and sequence-labeling-based~\cite{lou2018disfluency, zayats2016disfluency}. Many types of research have used sequence labeling models based on the begin-inside-outside (BIO) method that labels words as being inside or outside of a reparandum word sequence. Conditional random fields (CRFs)~\cite{ostendorf2013sequential, zayats2014multi}, long short-term memory (LSTM)~\cite{zayats2016disfluency}, and  auto-correlational neural network (ACNN)~\cite{lou2018disfluency} leverage the sequence labeling approach. \cite{bach2019noisy} demonstrated that the combination of BiLSTM, self-attention, and adding noise during training helps to achieve a performance comparable to that of the BERT method~\cite{devlin2018bert}. \cite{wang2019multi} proposed two multi-task self-supervised learning methods to tackle the bottleneck of training data and achieved comparable performance by using less than 1\% of the training data.

%% file: sections/3_model.tex
\section{Proposed Method}
\label{sec:method}
\subsection{Problem Definition}
Let \(\Theta_{s}\) represent the shared features of disfluency detection, NER, and POS tasks. Also, let \(\Theta_{d}\), \(\Theta_{e}\), and \(\Theta_{p}\) represent task specific features for disfluency detection, NER, and POS. We can now define disfluency detection as a sequence-labeling problem. Given a sequence \( X = \{x_{1}, x_{2},  \cdot \cdot \cdot, x_{n}\}\) where \( n\) denotes the number of tokens, and its corresponding disfluency labels \( Y_{d} = \{y_{1}, y_{2},  \cdot \cdot \cdot, y_{n}\}\), the disfluency detection task aims to learn the function \(F^{d}(\Theta_{s}, \Theta_{d}): X  \rightarrow Y_{d}\). 

Similarly, we define NER as a sequence-labeling problem, where given the same sequence \( X\) and its entity labels \( Y_{e} = \{y_{1}, y_{2},  \cdot \cdot \cdot, y_{n}\}\), learn the function \(F^{e}(\Theta_{s}, \Theta_{e}): X  \rightarrow Y_{e}\). Likewise, POS can be defined to learn the function \(F^{p}(\Theta_{s}, \Theta_{p}): X  \rightarrow Y_{p}\). Our objective is jointly training \(F^{d}(\Theta_{s}, \Theta_{d})\), \(F^{e}(\Theta_{s}, \Theta_{e})\), and \(F^{p}(\Theta_{s}, \Theta_{p})\) to learn the better shared features \(\Theta_{s}\).

\subsection{Contextual Representation}
Recently, the transformer model, which has an encoder and decoder, was proposed~\cite{vaswani2017attention}. Without recurrent layers, the transformer can encode contextual information using only self-attention mechanisms.  Based on the transformer architecture, pre-trained language models such as BERT~\cite{devlin2018bert} and ELECTRA~\cite{clark2020electra} have been proposed achieving state-of-the-art results on various NLP tasks.

In this work, we build our encoder based on transformer, BERT, and ELECTRA models. The sequence of token representation \( W=[w_{1},...,w_{n}]\) is fed into the encoder. Then the contextual representation \(H=[h_{1},...,h_{n}]\)) is obtained representing the history of context.

\subsection{Labeling with Conditional Random Field Decoder}
\label{sec:crf}

While the contextual representation generated by the encoder can take into account the attention between inputs, considering the information of neighboring labels is also important. A conditional random field (CRF) can consider the state transition probability between neighboring labels decoding the most probable output label sequence \cite{lafferty2001conditional}. Therefore, we adopted a CRF on top of the encoder to label disfluencies, named entities, and part-of-speech tags efficiently. The objective of the CRF is to maximize the log probability of the ground label \( \log p(Y|H) \), where H denotes the contextual representation and Y denotes the sequence of ground labels.

\subsection{Joint Training Objective}
Using three CRF decoders, we define the negative log likelihood loss for disfluency detection as \(\mathcal{L}_{d} = - \sum \log p(y^{d}_{t}|h_{t})\), NER as \(\mathcal{L}_{e} = - \sum \log p(y^{e}_{t}|h_{t})\), and POS as \(\mathcal{L}_{p} = - \sum \log\) \(p(y^{p}_{t}|h_{t})\), where \( \mathcal{L}_{d}\) denotes the CRF loss for disfluency detection, \( L_{e}\)  denotes the NER loss, and \( L_{p}\) denotes the POS loss. To utilize NER and POS as auxiliary tasks, we define the joint training objective for disfluency detection as \(\mathcal{L} = \mathcal{L}_{d} + \alpha (\mathcal{L}_{e} + \mathcal{L}_{p}) \), where \( \alpha \)
is coefficients determining the degree of NER and POS loss. Each encoder has different performances according to the value of \(\alpha \), and the results are described in the next section (see Figure~\ref{fig:f1_dev_results}). Note that we utilize NER and POS as auxiliary tasks in training time. Therefore, after training is completed, we use only one CRF layer for disfluency detection during inference time.

%% file: sections/4_experiment.tex
\section{Experiments}
\label{sec:experiments}
\begin{figure*}[t]\centering
\includegraphics[width=0.8\textwidth]{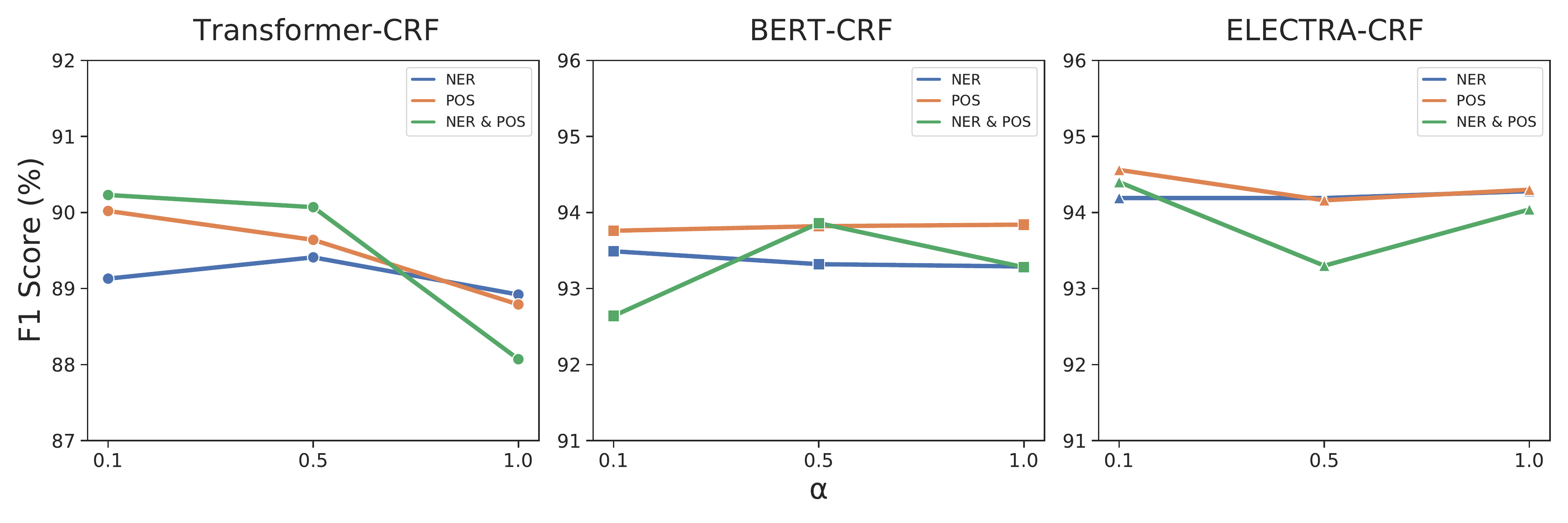}
\caption{F-scores of each model depending on the value of \(\alpha\) for each auxiliary SL task on the Switchboard dev dataset.}
\label{fig:f1_dev_results}
\vspace{-0.3cm}
\end{figure*}
\subsection{Dataset and Training Setup}
\noindent \textbf{Switchboard} \indent The English Switchboard dataset is the most widely used benchmark dataset for disfluency detection~\cite{godfrey1993switchboard}. It is consists of conversational speeches and annotated as in Figure~\ref{fig:example_1}. Following \cite{lou2018disfluency}, we defined disfluency detection as a binary classification problem, where reparandum are annotated as disfluency, while all other words are annotated as fluent speech. We use sw[23]*.dff files as training, sw4[5-9]*.dff files as validation, and sw4[0-1]*.dff files as test dataset following the experiment settings as in \cite{charniak2001edit}.

We use the Flair~\cite{akbik2019naacl, akbik2019flair, akbik2018coling} NER model~\footnote{https://github.com/flairNLP/flair} trained on the CoNLL NER dataset to assign named entities in a training data. Flair NER model was trained to predict 4 entities (e.g., `Locations (LOC)', `miscellaneous (MISC)', `organizations (ORG)', and `persons (PER)'). Also, the POS tag is labeled on the Switchboard dataset, and we used it to train the POS prediction task.

\subsection{Baselines}
To demonstrate the effectiveness of the proposed method, we build three baseline systems depending on which model to use as an encoder. We use the CRF as the decoder as described in~\ref{sec:crf}.

\noindent \textbf{Transformer-CRF} \indent The Transformer is an attention-based model~\cite{vaswani2017attention}. The Transformer encoder layer has consisted of a multi-head self-attention and a feed-forward sub-layer.

\noindent \textbf{BERT-CRF} \indent The BERT is a language model based on Transformer, and trained on a large-scale corpus with masked language model and next sentence prediction objectives~\cite{devlin2018bert}. 

\noindent \textbf{ELECTRA-CRF} \indent The ELECTRA is also a language model like BERT but trained with more effective pre-training method~\cite{clark2020electra}. In ELECTRA, instead of replacing some potion of token with [MASK] which corrupt inputs, replace some tokens with sampled from a generator. Then, a discriminator is trained to predict whether a generator replaces each token or not.

\subsection{Implementation Details}
We implement our model using the PyTorch \cite{adam2019pytorch} deep learning library. Specifically, we set the number of a layer as 2 and the number of a head as 8 for the Transformer. For language models, we adopt initial checkpoints for BERT and ELECTRA models from previous works~\cite{devlin2018bert,clark2020electra} and \textit{base} pre-trained language models from huggingface open source~\cite{wolf2020transformers}. For fine-tuning on the Switchboard dataset, we trained the models with a batch size of 32 using adam optimizer with an initial learning rate of 5e-5. All experiments are run on 4 Tesla V100 GPUs. Our code will be released upon acceptance to facilitate further researches.

\subsection{Evaluation Metrics}
We use token-based precision (P), recall (R), and F-score (F1) as the evaluation metrics following previous works~\cite{wang2017transition, wang2019multi}. Since the performance may vary due to different initialization values for each experiment, we report averaged scores across five experiments of each model.

\section{Results and Discussion}
\subsection{Performance on the English Switchboard Dataset}
\input{tables/tab1_test_results}

In our auxiliary sequence-labeling (SL) tasks, NER and POS tasks are jointly trained with disfluency detection task. The \(\alpha\) of each model is chosen by best-achieved F-score on the dev dataset. Figure~\ref{fig:f1_dev_results} reports F-scores of each model depending on the value of \(\alpha\) for each auxiliary SL task on the Switchboard dev dataset. As a result, we set the value of \(\alpha\) to 0.1 for Transformer-CRF, 0.5 for BERT-CRF, and 0.1 for ELECTRA-CRF. 

\input{tables/tab3_qual}

Table~\ref{tab:test_results} shows the evaluation results compared to the previous works on the Switchboard test dataset. First, we can observe that our BERT-CRF model with auxiliary SL tasks outperforms previous state-of-the-art model. Second, ELECTRA-CRF shows higher performance than BERT-CRF by 0.2 F-scores. As a result, we achieve new state-of-the-art result over previous works. Note that we do not compare our work with~\cite{wang2018semi}, since~\cite{wang2018semi} tagged RM and IM as disfluency, while other works only tagged RM as disfluency, including ours. As described in Table~\ref{table:speed}, since we utilize auxiliary SL tasks only in training time, the inference speeds are the same between the model with auxiliary SL tasks and the model without SL tasks.

\input{tables/tab4_speed}

\subsection{Ablation Analysis}
\label{sec:ablation}
\input{tables/tab2_abalations}

We conduct ablation analysis to investigate which auxiliary SL tasks are influential. Table~\ref{tab:ablations} reports ablation analysis on the Switchboard test dataset. In the case of Transformer-CRF, when using the NER task together compared to not using any SL tasks, the F-score is 1.1\% higher, and when POS is used, 1.6\% higher. Furthermore, when NER and POS tasks are used together, the F-score is 1.8\% higher. BERT-CRF shows 0.1\% higher score when using the NER task, 0.7\% higher score when using the POS task, and 0.7\% higher score when using NER and POS task are used together. Finally, ELECTRA-CRF shows 0.1\% higher score when using the NER task, 0.4\% higher score when using the POS task, and 0.4\% higher score when using NER and POS task are used together. Based on these results, the use of POS task in all models shows higher performance improvement than NER, and in the case of Transformer-CRF, using both NER and POS shows the highest performance improvement. Although the marginal improvement was achieved in the ELECTRA-CRF model, we believe that this is because the pre-trained ELECTRA model learns some of the named entities and POS information during the pre-training procedure. However, bigger improvements in Transformer and BERT-based models are more feasible in a real-world application in terms of cost-efficiency. Also, considering that there is not much difference in performance between previous state-of-the-art models in this task, we believe that absolute improvements of 1.8\%, 0.7\%, and 0.4\% for Transformer-CRF, BERT-CRF and ELECTRA-CRF are significant, which are gained from jointly training auxiliary SL tasks.

\subsection{Qualitative Analysis}
We also conduct a qualitative analysis on the English Switchboard test dataset. E1 and E2 represent examples of sentences containing named entities, and E3 and E4 represent POS tags of words corresponding to disfluency. Both in E1 and E2, the model trained with proposed auxiliary SL tasks correctly detects disfluencies while the model trained without auxiliary SL tasks makes inaccurate predictions. Specifically, in E1, the model without w/o aux predicts mceneil as a disfluency (false positive). But mceneil is the name of a person while neil- is not a certain entity which is a ground-truth disfluency. Likewise, in E2, ``the eight hours" represents the time entity and the model w/o aux inaccurately predicts it as a disfluency. In E3, the same POS tag sequence is used repeatedly, as describe in Figure~\ref{fig:example_2} and the model with auxiliary SL tasks correctly detects disfluencies. Also, in E4, ground-truth disfluencies are ``we" and ``," having POS tags ``$<$prp$>$" and ``$<$,$>$". As described in the introduction, the ratio of these two POS tags constituting disfluencies accounts for 55\%, and our model with auxiliary SL tasks correctly detects disfluencies while the model w/o aux is not. These observations suggest that jointly training NER and POS tasks with disfluency detection task is highly effective by allowing the shared encoder to learn a deep semantic representation of words reflecting NER and POS meaning representation.

%% file: tables/tab1_test_results.tex
\begin{table}[h]\centering
\begin{adjustbox}{width=0.43\textwidth}
\renewcommand{\arraystretch}{1.0}%
\begin{tabular}{lccc}
\hline
Model            & P    & R    & F1   \\ \hline
Semi-CRF~\cite{ferguson2015disfluency} & 90.0 & 81.2 & 85.4 \\
Bi-LSTM~\cite{zayats2016disfluency}          & 91.6 & 80.3 & 85.9 \\
Attention-based~\cite{wang2016neural}  & 91.6 & 82.3 & 86.7 \\
Transition-based~\cite{wang2017transition} & 91.1 & 84.1 & 87.5 \\
Self-supervised ~\cite{wang2019multi} & 93.4 & 87.3 & 90.2 \\
Self-trained~\cite{lou2020improving}& 87.5 & \textbf{93.8} & 90.6 \\
EGBC~\cite{bach2019noisy}             & \textbf{95.7} & 88.3 & 91.8 \\ 
BERT fine-tune~\cite{bach2019noisy}             & 94.7 & 89.8 & 92.2
\\ \hline
Auxiliary SL Tasks (Ours) \\ \hline
Transformer-CRF & 93.3 & 84.8 & 89.2 \\ 
BERT-CRF        & 94.6 & 91.2 & \underline{92.9} \\ 
ELECTRA-CRF     & \underline{94.8} & \underline{91.6} & \textbf{93.1} \\ 

\hline
\end{tabular}
\end{adjustbox}
\caption{Evaluation results compared to the existing models on the Switchboard test dataset. The best scores are in bold, and second best scores are underlined.}
\label{tab:test_results}
\vspace{-0.5cm}
\end{table}

%% file: tables/tab3_qual.tex
\begin{table*}[t]\centering 
\centering
\resizebox{1.0\textwidth}{!}{
\begin{tabular}{clccc}
\toprule

& Examples   & Model & Predictions   \\
\midrule
\multirow{2}{*}{E1} & \multirow{2}{*}{\begin{tabular}[c]{@{}l@{}}well, mcneil\textcolor{darkgreen}{$($B-PERSON$)$}, \textcolor{medblue}{neil-}, and lehrer\textcolor{darkgreen}{$($B-PERSON$)$}.\end{tabular}}
      & Ours w/o aux  & mcneil,   \\
      \cdashline{3-5}
&     & Ours  & neil-,   \\

\midrule
\multirow{2}{*}{E2} & \multirow{2}{*}{\begin{tabular}[c]{@{}l@{}}i mean, the\textcolor{darkgreen}{$($B-TIME$)$} eight\textcolor{darkgreen}{$($I$)$} hours\textcolor{darkgreen}{$($I$)$}, during the day, when they 're supposed to be there, \\i think they have every right to say \textcolor{medblue}{this is}, these behaviors are acceptable and these are not.\end{tabular}}
      & Ours w/o aux & the eight hours,   \\
      \cdashline{3-5}
&     & Ours  & this is,   \\

\midrule
\multirow{2}{*}{E3} & \multirow{2}{*}{\begin{tabular}[c]{@{}l@{}}i think that \textcolor{medblue}{they}\textcolor{darkred}{$<$prp$>$} \textcolor{medblue}{'re}\textcolor{darkred}{$<$vbp$>$} they\textcolor{darkred}{$<$prp$>$} 're\textcolor{darkred}{$<$vbp$>$} operating on a more,  \end{tabular}}
      & Ours w/o aux  & -   \\
      \cdashline{3-5}
&     & Ours  & they 're   \\

\midrule
\multirow{2}{*}{E4} & \multirow{2}{*}{\begin{tabular}[c]{@{}l@{}}\textcolor{medblue}{we}\textcolor{darkred}{$<$prp$>$} \textcolor{medblue}{,}\textcolor{darkred}{$<$,$>$} i don't think we have as much of the gang problem as a lot of the other cities have. \end{tabular}}
      & Ours w/o aux  & -   \\
      \cdashline{3-5}
&     & Ours  & we ,   \\
     
\bottomrule
\end{tabular}
}
\caption{Case study on the Switchboard test dataset. Words painted in blue represent ground-truth disfluencies. Named entities are painted in green, and part-of-speech tags are painted in red. Ours w/o aux denotes the model trained without auxiliary SL tasks. All models are built based on the Transformer-CRF model. `-' denotes that the model failed to extract corresponding disfluencies.}
\label{table:qual}
\vspace{-0.3cm}

\end{table*}

%% file: tables/tab4_speed.tex
\begin{table}[h]\centering
\begin{adjustbox}{width=0.33\textwidth}
\renewcommand{\arraystretch}{1.2}%
\centering
\begin{tabular}{ll}
\hline
\textbf{Model} & \textbf{number of sentences per second} \\ \hline
Ours w/o aux      & 324    \\ 
Ours     & 324    \\ \hline
\end{tabular}
\end{adjustbox}
\caption{Comparison of inference speed between Ours w/o aux and Ours with auxiliary SL tasks. Ours w/o aux denotes ours without auxiliary SL tasks. The batch size is set to 1 considering the single prediction environment with real-time. All models are based on the Transformer-CRF model and results are the average of 10 experiments.}
\label{table:speed}
\vspace{-0.3cm}
\end{table}

%% file: tables/tab2_abalations.tex
\begin{table}[h]\centering
\begin{adjustbox}{width=0.35\textwidth}
\renewcommand{\arraystretch}{0.95}%
\begin{tabular}{lccc}
\hline
Model            & P    & R    & F1   \\ \hline
Transformer-CRF     & 92.3 & 83.0 & 87.4 \\ 
\;\;+ NER           & 91.6 & \textbf{85.6} & 88.5$^{\dagger}$ \\ 
\;\;+ POS           & 92.8 & 85.4 & 89.0$^{\dagger}$ \\
\;\;+ NER + POS     & \textbf{93.3} & 84.8 & \textbf{89.2}$^{\dagger}$ \\ 
\hline
BERT-CRF            & 94.5 & 90.0 & 92.2 \\
\;\;+ NER           & 95.4 & 89.4 & 92.3$^{\dagger}$ \\ 
\;\;+ POS           & 94.6 & \textbf{91.3} & 92.9$^{\dagger}$ \\ 
\;\;+ NER + POS     & \textbf{94.6} & 91.2 & \textbf{92.9}$^{\dagger}$ \\ 
\hline
ELECTRA-CRF         & 95.3 & 90.1 & 92.7 \\ 
\;\;+ NER           & \textbf{95.8} & 90.0 & 92.8$^{\dagger}$ \\ 
\;\;+ POS           & 94.1 & \textbf{92.1} & 93.1$^{\dagger}$ \\ 
\;\;+ NER + POS     & 94.8 & 91.6 & \textbf{93.1}$^{\dagger}$ \\ 
\hline
\end{tabular}
\end{adjustbox}
\caption{Ablation analysis on the Switchboard test dataset. $\dagger$ denotes statistical significance (p-value $<$ 0.05).}
\label{tab:ablations}
\vspace{-0.5cm}
\end{table}

%% file: sections/5_conclusion.tex
\section{Conclusion}
\label{sec:print}
In this paper, we proposed the joint training method utilizing NER and POS as auxiliary tasks in disfluency detection. Through extensive evaluations, we showed that on the widely used English Switchboard dataset, the joint training method could lead to achieving better f-score and achieved state-of-the-art result. In future work, we will investigate whether auxiliary tasks are helpful in disfluency detection in other languages as well.